\documentclass[conference]{IEEEtran}
\IEEEoverridecommandlockouts

\usepackage{amsmath,amssymb,amsfonts}
\usepackage{algorithm}
\usepackage{graphicx}
\usepackage{subcaption}
\usepackage{textcomp}
\usepackage{xcolor}
\usepackage{biblatex}
\usepackage{algpseudocode}
\usepackage[font=footnotesize]{caption}

\newcommand{\bphi}{\boldsymbol{\phi}}
\newcommand{\norm}[1]{\left\Arrowvert #1 \right\Arrowvert}

\newcommand{\soft}[1]{\textsc{#1}}

\newcommand{\name}{DDM-GNN}
\newcommand{\ddmlu}{DDM-LU}
\newcommand{\dsstheta}{\text{DSS}_\theta}

\bibliography{biblio_file.bib}

\def\BibTeX{{\rm B\kern-.05em{\sc i\kern-.025em b}\kern-.08em
    T\kern-.1667em\lower.7ex\hbox{E}\kern-.125emX}}

\begin{document}


\title{Multi-Level GNN Preconditioner for Solving \\ Large Scale Problems \\
}

\author{\IEEEauthorblockN{Matthieu Nastorg}
\IEEEauthorblockA{\textit{Université Paris-Saclay} \\
\textit{CNRS, Inria, LISN}\\
91405 Orsay, France \\
matthieu.nastorg@inria.fr}
\and
\IEEEauthorblockN{Jean-Marc Gratien}
\IEEEauthorblockA{\textit{Direction Technologie et Numerique} \\
\textit{IFP Énergies-Nouvelles}\\
92852 Rueil-Malmaison, France}
\and
\IEEEauthorblockN{Thibault Faney}
\IEEEauthorblockA{\textit{Direction Technologie et Numerique} \\
\textit{IFP Énergies-Nouvelles}\\
92852 Rueil-Malmaison, France}
\and
\IEEEauthorblockN{Michele Alessandro Bucci}
\IEEEauthorblockA{\textit{Digital Sciences \& Technologies Department} \\
\textit{Safran Tech}\\
78114 Châteaufort, France}
\and
\IEEEauthorblockN{Guillaume Charpiat}
\IEEEauthorblockA{\textit{Université Paris-Saclay} \\
\textit{CNRS, Inria, LISN}\\
91405 Orsay, France}
\and
\IEEEauthorblockN{Marc Schoenauer}
\IEEEauthorblockA{\textit{Université Paris-Saclay} \\
\textit{CNRS, Inria, LISN}\\
91405 Orsay, France}
}

\maketitle

\begin{abstract} 
Large-scale numerical simulations often come at the expense of daunting computations. High-Performance Computing has enhanced the process, but adapting legacy codes to leverage parallel GPU computations remains challenging. Meanwhile, Machine Learning models can harness GPU computations effectively but often struggle with generalization and accuracy. Graph Neural Networks (GNNs), in particular, are great for learning from unstructured data like meshes but are often limited to small-scale problems. Moreover, the capabilities of the trained model usually restrict the accuracy of the data-driven solution. To benefit from both worlds, this paper introduces a novel preconditioner integrating a GNN model within a multi-level Domain Decomposition framework. The proposed GNN-based preconditioner is used to enhance the efficiency of a Krylov method, resulting in a hybrid solver that can converge with any desired level of accuracy. The efficiency of the Krylov method greatly benefits from the GNN preconditioner, which is adaptable to meshes of any size and shape, is executed on GPUs, and features a multi-level approach to enforce the scalability of the entire process. Several experiments are conducted to validate the numerical behavior of the hybrid solver, and an in-depth analysis of its performance is proposed to assess its competitiveness against a C++ legacy solver.
\end{abstract}


\section{Introduction}

Partial Differential Equations (PDEs) are essential for understanding complex physical or artificial processes in science and engineering. However, solving these equations at scale can be challenging due to the computational cost of resolving the smallest spatio-temporal scales. The steady-state Poisson equation is one of the most common and widely used PDEs. It is ubiquitous \cite{pressure_poisson, gravity_poisson, electrostatics_poisson, surface_reconstruction_poisson}, and plays a critical role in modern numerical solvers.
Despite significant progress in the High-Performance Computing (HPC) community, the solution to the Poisson problem remains a major bottleneck in many numerical simulation processes \cite{phd_yushan_wang}. \\

Nowadays, data-driven methods like Deep Neural Networks (DNNs) are reshaping the field of numerical simulations. These deep networks can offer faster predictions, thereby reducing turnaround time for workflows in engineering and science \cite{latent_space_physics, ml_accelerated_cfd}. Early attempts to apply DNNs to solve PDEs involved treating simulation data as images and leveraging Convolutional Neural Networks (CNNs) \cite{ozbay_poisson_2021, illarramendi2022performance}. CNN models have demonstrated remarkable performance in processing structured, image-like data. However, their effectiveness is limited when dealing with unstructured data, such as meshes encountered in numerical simulations. To overcome this limitation, researchers have turned to Graph Neural Networks (GNNs), which have displayed great potential in solving PDEs on unstructured grids \cite{pfaff_2020_learning, horie_2022_physics}.
Using data-driven methods to solve PDEs offers several benefits compared to traditional solvers \cite{iterative_methods_sparse_linear_systems}. Machine Learning (ML) models usually function as ``black-box'' approaches, approximating the solution without the need for the costly computations of creating and solving the system of equations. Data-driven methods are also well-suited for parallel computations on GPUs, unlike traditional methods that rely on CPU computations with limited parallelization on multiple cores. Conversely, ML methods can suffer from various issues, such as:\\

\textbf{Generalization} \quad ML models can provide approximate solutions to problems within a specific training distribution, but out-of-distribution samples are often poorly solved. For instance, GNN models for PDEs are usually constrained to small-scale problems with fixed-size meshes, hindering their practical use in industrial contexts.\\

\textbf{Level of Accuracy} \quad The precision of an ML model is limited. It is as precise as the capabilities of the trained model permit and often deteriorates further when handling out-of-distribution samples. When a very accurate PDE solution is needed, leveraging an ML model can quickly be ill-suited (e.g., accurately solving a Poisson Pressure problem to ensure the consistency of a fractional step method \cite{pressure_poisson}).

In short, traditional solvers guarantee convergence but are limited to CPU computations. On the other hand, data-driven methods harness the full power of parallel GPU computations but lack generalization and convergence guarantees. To address the generalization capabilities of ML models, a potential solution involves building hybrid solvers that combine Machine Learning with Domain Decomposition Methods (DDMs), which is a recent and promising field \cite{ml_ddm_survey_1, ml_ddm_survey_2}. Some research has focused on using ML models to enhance DDMs by approximating optimal interface conditions \cite{learn_interface_1, learn_interface_2}. Other work involves using different types of DNNs to replace the local sub-domain or coarse solvers. In \cite{deep_ddm} and \cite{d3m}, the sub-domain solvers in a parallel overlapping Schwarz method are replaced by PINNs \cite{pinns} and the Deep Ritz method \cite{deep_ritz}, respectively. 
While these methods yield satisfactory outcomes, they are typically limited to solving PDEs on small-scale Cartesian data. Extending them for large-scale problems with varying shapes and sizes remains elusive, as does fair evaluation of their performance in real industrial contexts. Besides, these approaches still operate as entire data-driven frameworks with no guarantee of converging to any desired precision. \\

We address these limitations with a novel GNN-based preconditioner, combining GNN solver techniques with multi-level Domain Decomposition. This innovative preconditioner mirrors the structure of an Additive Schwarz Method \cite{history_schwarz_method}, with GNN models tackling multiple sub-problems. This flexibility allows efficient handling of large-scale meshes with varying sizes and shapes by adapting sub-problem sizes to GNN capabilities. Leveraging GPU parallelism, the method concurrently solves multiple sub-problems in batches. This results in a hybrid solver that can achieve high precision convergence through Krylov methods while significantly boosting their efficiency thanks to the GNN preconditioner, efficiently executed on GPUs. Additionally, the preconditioner can incorporate a two-level method to enhance weak scalability regarding the number of sub-domains.\\

In this paper, Section \ref{sec:background} first reviews the main principles of the algorithms used in the fields of Domain Decomposition and statistical GNN-based solvers. Then, Section \ref{sec:methodology} introduces the hybrid methodology, which combines algorithms from the two fields to overcome their limitations. Section \ref{sec:results} showcases the results validating that our approach is numerically accurate. The performance of our method is benchmarked against that of an optimized C++ legacy linear solver to determine if the methodology is competitive with a state-of-the-art solver used in real CFD software. Finally, we conclude our work and provide perspectives for future research.

\section{Background}
\label{sec:background}

This section provides background materials, reviewing the multi-level Additive Schwarz Preconditioner in Section \ref{subsec:asm}, and Graph Neural Networks in Section \ref{subsec:dss}. Let us first state the problem addressed in the entire paper.\\

Let $\Omega \subset \mathbb{R}^n$ be a bounded open domain with smooth boundary $\partial \Omega$. Let $f$ be a continuous function defined on $\Omega$, and $g$ a continuous function defined on $\partial \Omega$. In this paper, we focus on solving Poisson problems with Dirichlet boundary conditions, which consists in finding a real-valued function $u$, defined on $\Omega$, solution of:

\begin{equation}
\left \{
\begin{array}{rcl}
-\Delta u &=& f \qquad \in \Omega \\
u &=& g \qquad \in \partial \Omega \\
\end{array}
\right.
\label{poisson_equation}
\end{equation}

Except in very specific instances, no analytical solution can be derived for the Poisson problem, and its solution must be numerically approximated: the domain $\Omega$ is first discretized into an unstructured mesh, denoted $\Omega_h$. The Poisson equation \eqref{poisson_equation} is then spatially discretized using the Finite Element Method. The approximate solution is sought as a vector of values defined on all $N$ degrees of freedom of $\Omega_h$. In this work, we use first-order Lagrange elements, thus $N$ matches the number of nodes in $\Omega_h$. The discretization of the variational formulation of \eqref{poisson_equation} using Galerkin's method results in solving a linear system of the form:

\begin{equation}
A\mathbf{u} = \mathbf{b}
\label{linear-system}
\end{equation} 

where the matrix $A \in \mathbb{R}^{N\times N}$ is sparse and represents the discretization of the continuous Laplace operator, the vector $\mathbf{b} \in \mathbb{R}^N$ comes from the discretization of the forcing term $f$ and of the boundary conditions $g$, and $\mathbf{u} \in \mathbb{R}^N$ is the solution vector to be sought. \\

For complex industrial problems, accurate predictions are mandatory, leading to an increasingly large $N$ and, thus, an extensive system \eqref{linear-system} to solve. 
Iterative solvers, particularly Krylov methods like Conjugate Gradient (CG), BiCGStab, and GMRES \cite{iterative_methods_sparse_linear_systems}, are favored for their superior convergence compared to stationary algorithms. However, they still face challenges related to efficiency and scalability. Efficiency is tied to the convergence rate, where a higher rate requires fewer iterations to achieve a desired precision, although this rate decreases with larger problem sizes due to matrix conditioning. Scalability concerns how solution time responds to increased computational resources, ideally remaining constant for a fixed problem size per core.

\subsection{Multi-level Additive Schwarz Preconditioner}
\label{subsec:asm}

Preconditioning can significantly improve the efficiency and scalability of Krylov methods. 
In practice, the rationale is to find a matrix $M \in \mathbb{R}^{N \times N}$ and to use a Krylov method to solve the following preconditioned problem:

\begin{equation}
    M^{-1}A\mathbf{u} = M^{-1}\mathbf{b}
    \label{preconditioned_system}
\end{equation}

Preconditioners are generally designed to improve the efficiency of iterative methods, but not all can enhance their scalability. In this context, we review the multi-level Additive Schwarz Method (ASM), which offers both properties. ASM belongs to Schwarz methods \cite{schwarz1870ueber} from the broad field of Domain Decomposition Methods (DDMs) \cite{dolean2015introduction}. DDMs leverage the principle of ``divide and conquer'': the global problem \eqref{poisson_equation} is partitioned into sub-problems of manageable size that can be solved in parallel on multiple processor cores. Initially introduced as stationary iterative methods, Schwarz methods are commonly applied as preconditioners for Krylov methods. Let us consider an overlapping decomposition of $\Omega$ into $K$ open sub-domains $(\Omega_i)_{1\leq i\leq K}$ such that: 
$$\Omega = \displaystyle \bigcup_{i=1}^K \Omega_i $$
Let $u^n$, $n \in \mathbb{N}$ be an approximate solution to the Poisson problem \eqref{poisson_equation}. As an iterative solver, the one-level ASM computes an updated approximate solution $u^{n+1}$ by first solving, for each sub-domain $i = 1, \dots, K$, the following local sub-problems:

\begin{equation}
    \left \{
    \begin{array}{rll}
    -\Delta v_i^n & = r^n & ~\text{in}~ \Omega_i\\
    v_i^n & = 0 & ~\text{on}~ \partial \Omega_i
    \end{array}
    \right.        
    \label{local_subproblems}
\end{equation}

where $r^n = f + \Delta u^n$ is the residual at iteration $n$. The updated approximation $u^{n+1}$ is then obtained using all sub-solutions $v_i^n$ such that:

\begin{equation}
    u^{n+1} = u^n + \sum_{i=1}^K E_i(v_i^n)
    \label{asm_updates}
\end{equation}

where $E_i$ is an extension operator that maps functions defined on $\Omega_i$ to their extension on $\Omega$ that takes value $0$ outside $\Omega_i$. \\

From a numerical perspective, a discrete counterpart can be established to solve the system \eqref{linear-system}. Let us consider now an overlapping decomposition of the mesh $\Omega_h$ into $K$ sub-meshes $(\Omega_{h,i})_{1\leq i\leq K}$ such that:

\begin{equation*}
    \Omega_h = \displaystyle \bigcup_{i=1}^K \Omega_{h,i}.
\end{equation*}

We denote by $k_i$ the number of nodes in the sub-mesh $i$. The restriction of a solution vector $\mathbf{u} \in \mathbb{R}^N$ to a sub-mesh $\Omega_{h,i}$ can be expressed as $R_i \mathbf{u}$, where $R_i$ is a rectangular boolean matrix of size $k_i \times N$. The extension operator is defined as the transpose matrix $R_i^T$. With these notations, the one-level ASM preconditioner, referred to as $M^{-1}_{\text{ASM},1}$, is defined as:

\begin{equation}
    M^{-1}_{\text{ASM},1} = \sum_{i=1}^K R_i^T(R_iAR_i^T)^{-1}R_i
    \label{one_level_preconditioner}
\end{equation}

The one-level ASM preconditioner \eqref{one_level_preconditioner} is not scalable regarding the number of sub-domains. 
One mechanism to enhance scalability involves implementing a two-level method with coarse space correction, which connects all sub-domains at each iteration.
A potential solution is to leverage a Nicolaides coarse space \cite{nicolaides1987deflation}, which defines a matrix $R_0$ of size $K \times N$. 
We refer the reader to \cite{dolean2015introduction} for additional information. The two-level Additive Schwarz preconditioner, referred to as $M^{-1}_{\text{ASM},2}$, is then defined as:

\begin{equation}
    M^{-1}_{\text{ASM},2} = \underbrace{R_0^T(R_0AR_0^T)^{-1}R_0}_{\text{coarse problem}} + \underbrace{\sum_{k=1}^K R_i^T(R_iAR_i^T)^{-1}R_i}_{\text{local problems}}
    \label{two_level_preconditioner}
\end{equation}

The continuous iterative formulation of ASM, defined by equations \eqref{local_subproblems} and \eqref{asm_updates}, can be expressed in a discrete form by the following preconditioned fixed-point iteration:

\begin{align}
\mathbf{u}^{n+1} & = \mathbf{u}^n + M^{-1}_{\text{ASM}} \left(\mathbf{b} - A\mathbf{u}^n\right) \\
& = \mathbf{u}^n + M^{-1}_{\text{ASM}}~\mathbf{r}^n
\label{asm_fp}
\end{align}

where $M^{-1}_{\text{ASM}}$ stands for the one or two level ASM preconditioner. Given its symmetric nature, the multi-level ASM preconditioner $M^{-1}_{\text{ASM}}$ performs remarkably well when used as a preconditioner for the Preconditioned Conjugate Gradient (PCG) algorithm \cite{dolean2015introduction}. The PCG algorithm is reminded in Algorithm \ref{algo:pcg}, including the application of the multi-level ASM preconditioner to residual vectors highlighted in red. \\

\begin{algorithm}[tb]
    \caption{Preconditioned Conjugate Gradient}
    \vspace{0.1cm}
    \begin{algorithmic}
    \State \textbf{Compute:} \\ 
    \vspace{0.05cm}
    $\mathbf{r}_0 = \mathbf{b} - A\mathbf{u}_0$, \quad $\mathbf{z}_0 = {\color{red}M^{-1}_{\text{ASM}}}\left(\mathbf{r}_0\right)$, \quad $\mathbf{p}_0 = z_0$
    \vspace{0.1cm}
    \For{$i = 0, 1, \dots$}
    \vspace{0.1cm}
        \State $\rho_i = \langle \mathbf{r}_i, \mathbf{z}_i \rangle$,
        \quad $\mathbf{q}_i = A\mathbf{p}_i$, \quad  $\alpha_i = \dfrac{\rho_i}{\langle \mathbf{p}_i, \mathbf{q}_i\rangle}$
        \State $\mathbf{u}_{i+1} = \mathbf{u}_i + \alpha_i\mathbf{p}_i$
        \State $\mathbf{r}_{i+1} = \mathbf{r}_i - \alpha_i\mathbf{q}_i$
        \vspace{0.1cm}
        \If{$\norm{\mathbf{r}_{i+1}} < \text{tol}$} \small Break;
        \EndIf
        \vspace{0.1cm}        
        \State $\mathbf{z}_{i+1} = {\color{red}M^{-1}_{\text{ASM}}}\left(\mathbf{r}_{i+1}\right)$, ~ $\rho_{i+1} = \langle \mathbf{r}_{i+1}, \mathbf{z}_{i+1} \rangle$, ~ $\beta_{i+1} = \dfrac{\rho_{i+1}}{\rho_i}$  
        \State $\mathbf{p}_{i+1} = \mathbf{z}_{i+1} + \beta_{i+1}\mathbf{p}_i$
    \vspace{0.1cm}
    \EndFor
    \end{algorithmic}
    \label{algo:pcg}
\end{algorithm}

\subsection{Deep Statistical Solvers}
\label{subsec:dss}

This section introduces Deep Statistical Solvers (DSS) \cite{deep_statistical_solver}, a promising GNN approach used in Machine Learning to solve Poisson problems like \eqref{poisson_equation} on unstructured meshes. In this study, the loss function is a ``physics-informed'' loss defined as the residual of the discretized Poisson problem. This approach differs from most previous ML methods, which aim to minimize the distance between the output of the model and a ``ground-truth'' solution. Therefore, a DSS model can be trained without a large set of expensive training sample solutions. Moreover, its physics-informed training process allows for better generalization when approaching problems on varying geometries compared to supervised methods. \\


The fundamental idea of the DSS approach is, considering a discrete Poisson problem like \eqref{linear-system}, to build a Machine Learning solver, parametrized by a vector $\theta$ and denoted as $\dsstheta$, which outputs an approximate solution $\tilde{\mathbf{u}}$ such that:



\begin{equation}
\tilde{\mathbf{u}} = \dsstheta\left(A,~\mathbf{b} \right) 
\end{equation}

Because the system \eqref{linear-system} was constructed using first-order finite elements, the matrix $A$ can be viewed as the adjacency matrix of its corresponding mesh $\Omega_h$. 
As a result, a discretized Poisson problem \eqref{linear-system} with $N$ degrees of freedom can be interpreted as a graph problem $G = (N,~A,~\mathbf{b})$, where $N$ is the number of nodes in the graph, $A = (a_{ij})_{(i,j) \in[N]^2}$ is the weighted adjacency matrix that represents the interactions between the nodes, and $\mathbf{b} = (b_i)_{i\in[N]}$ is some local external input applied to all the nodes in the graph. Vector $\mathbf{u} = (u_i)_{i\in[N]}$ represents the state of the graph, with $u_i$ being the state of node $i$. In addition, we define $\mathcal{L}_\text{res}$ as the real-valued function that computes the mean squared error of the discretized residual equation:

\begin{align}
\mathcal{L}_\text{res}(\mathbf{u},~G) & = \displaystyle \frac{1}{N}\sum_{i \in [N]} \Big( -b_i + \sum_{j \in [N]} a_{i,j}u_j \Big) ^2
\label{loss-function}
\end{align}

Let $\mathcal{S}$ and $\mathcal{U}$ be the set of all such graphs $G$ and all states $\mathbf{u}$ that belong to the same distribution of discretized Poisson problems. In \cite{deep_statistical_solver}, the statistical objective is, given a graph $G$ in $\mathcal{S}$, to find an optimal state in $\mathcal{U}$ that minimizes \eqref{loss-function}. Therefore, $\dsstheta$ is trained to predict from $G$ a solution $U$ to solve the following statistical problem: \\

\textit{Given a distribution $\mathcal{D}$ on space $\mathcal{S}$ and the loss $\mathcal{L}_\text{res}$, solve:}
\begin{equation}
\displaystyle \widehat{\theta} = \underset{{\theta}}{\text{argmin}}~\underset{G \sim \mathcal{D}}{\mathbb{E}} \left[\mathcal{L}_\text{res}\left(\text{DSS}_{\theta}\left(G\right),G\right)\right]
\label{ssp-problem}
\end{equation}

$\dsstheta$ uses an iterative architecture, propagating information through the mesh thanks to a manually set number of different Message Passing Neural Network (MPNN) layers. Interestingly, \cite{deep_statistical_solver} demonstrates, under mild hypotheses, the consistency of the approach. However, it depends on the number of MPNN layers, which should be directly proportional to the diameter of the meshes at hand. While DSS yields efficient results, it is limited to meshes of the same sizes due to its fixed number of MPNN layers. To tackle meshes at a large scale, the number of iterations must be set accordingly, leading to a significant increase in the model size, eventually becoming intractable.  \\

\section{Methodology}
\label{sec:methodology}

\begin{figure*}[tb]
\centering
\includegraphics[width=\textwidth]{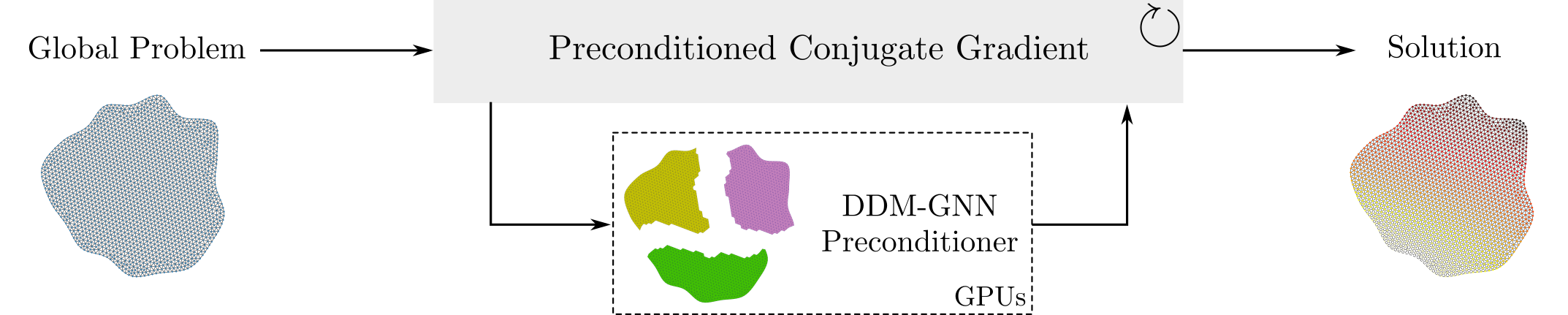}
\caption{Illustration of the proposed hybrid solver: A global Poisson problem (left) is solved using a Preconditioned Conjugate Gradient (PCG) method, enhanced by the \name{} preconditioner. \name{} combines Domain Decomposition with a GNN-based solver, harnessing parallel computations on GPUs.}
\label{fig:hybrid_solver}
\end{figure*}

Given the limitations outlined in Section \ref{subsec:dss}, the Deep Statistical Solvers method cannot effectively handle Poisson problems on a large scale. Instead, we propose using a Conjugate Gradient (CG) method preconditioned with a multi-level GNN preconditioner, referred to as \name\footnote{Domain Decomposition Method - Graph Neural Network}. \name{} adopts an Additive Schwarz approach (as described in Section \ref{subsec:asm}), where the resolution of the local problems \eqref{local_subproblems} is tackled using a DSS model. This approach aims to leverage both traditional and data-driven methods to solve Poisson problems like \eqref{poisson_equation} at scale. The proposed hybrid solver combines the convergence guarantee provided by the CG method with enhanced efficiency through the use of the \name{} preconditioner. \name, in turn, can effectively handle problems of any size and shape by choosing the size of the sub-problems in line with the capabilities of the DSS model. Additionally, \name{} harnesses GPU parallel computations to further enhance its performance and can be equipped with a two-level method to enable the scalability of the entire process. The resolution of a Poisson problem using the proposed hybrid solver is illustrated in Fig.\ref{fig:hybrid_solver}, and the code is available here\footnote{code will be made public after acceptance.}.\\

Section \ref{subsec:pipeline} provides a comprehensive explanation of the \name{} preconditioner, followed by a detailed description of the architecture of the DSS model in Section \ref{subsec:archi}.

\subsection{\name{} Preconditioner}
\label{subsec:pipeline}

The \name{} preconditioner is intended to replace the ASM preconditioner in Algorithm \ref{algo:pcg}. Following this setup, \name{} should take a global residual vector $\mathbf{r} \in \mathbb{R}^N$ as input, and output a global correction vector $\mathbf{z} \in \mathbb{R}^N$, ensuring the consistency of the PCG algorithm. \\

In ASM (Section \ref{subsec:asm}), the local problems \eqref{local_subproblems} differ from the global one \eqref{poisson_equation}. Despite this difference, there are still Poisson problems with Dirichlet boundary conditions, and using a DSS framework (Section \ref{subsec:dss}) to solve these local problems remains consistent. Notably, the force function $f$ in \eqref{poisson_equation} now corresponds to residuals $r$. It is noteworthy that the Dirichlet boundary conditions of these local problems are of a homogeneous type instead of being represented by some function $g$ in \eqref{poisson_equation}, significantly simplifying the distribution of problems addressed by the DSS model. \\

As the PCG algorithm aims to minimize the residual with each iteration, the source terms of local Poisson problems (i.e., residual vectors $\mathbf{r}$ in PCG) have increasingly smaller norms. This poses a challenge to our methodology: as the residual norm approaches zero, the DSS model might struggle to generalize properly since $\mathbf{r}$ serves as an input to the model. This could result in a trivial solution, i.e., a solution equal to zero everywhere. Consequently, PCG might stagnate at a certain precision threshold, as \name{} preconditioner would consistently provide a null correction to update the solution. To address this issue, we propose normalizing the source term of the local Poisson problems before providing it as input to the model. Formally, and using the same notations as defined in Section \ref{sec:background}, \name{} outputs a correction vector $\mathbf{z}$ in three steps, as follows:\\

\textbf{1. Resolution of the coarse problem} \quad The resolution of the coarse problem is made similarly to Section \ref{subsec:asm}, using LU decomposition to solve:

\begin{equation}
\mathbf{r}_c = R_0^T(R_0AR_0^T)^{-1}R_0\mathbf{r}
\end{equation}

\textbf{2. Resolution of local problems} \quad The $K$ local problems are solved concurrently on GPUs using a DSS model such that:

\begin{equation}
\left[\tilde{\mathbf{r}}_1, \dots, \tilde{\mathbf{r}}_K\right] = \dsstheta \left(\left[G_1, \dots , G_K\right]\right)
\label{eq:dss_asm_gnn}
\end{equation}

where 

\begin{equation}
G_i = \left(k_i, ~R_iAR_i^T, ~\frac{R_i\mathbf{r}}{||R_i\mathbf{r}||}\right)
\label{graph_solutions}
\end{equation}

represents the $i$-th graph associated to the $i$-th discretized local Poisson problem.
Here, \eqref{eq:dss_asm_gnn} suggests that all subdomains are solved simultaneously in one inference of $\dsstheta$. However, if the number of local problems becomes too large, $\left[G_1, \dots , G_K\right]$ can be partitioned into $N_b$ batches, allowing all problems to be solved in $N_b$ inferences of $\dsstheta$. \\

\textbf{3. Gluing everything together} \quad The correction vector $\mathbf{z}$ is finally computed by combining the extended coarse problem with the extended local solutions such that:

\begin{equation}
\mathbf{z} = \mathbf{r}_c + \sum_{i=1}^{K} R_i^T~||R_i\mathbf{r}||~\tilde{\mathbf{r}}_i
\end{equation}

\begin{figure}[b]
  \includegraphics[width=\columnwidth]{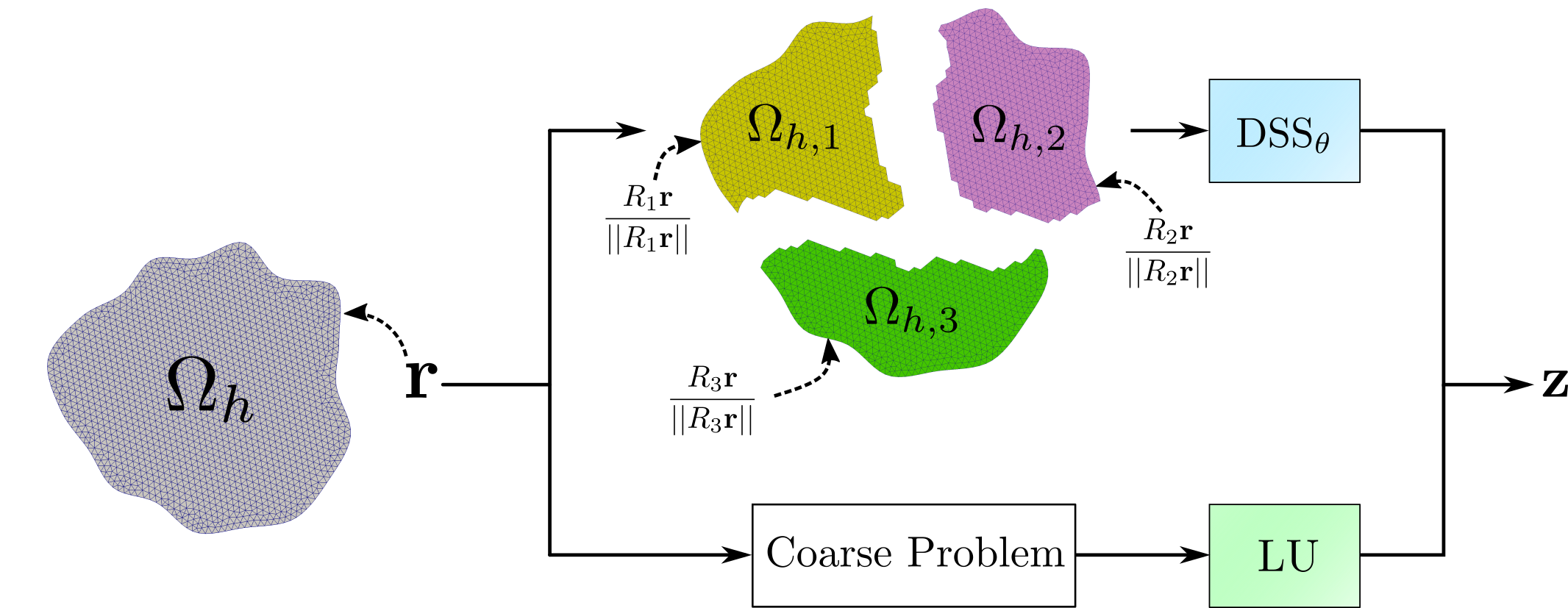}
  \caption{Illustration of \name{}: It takes a global residual vector $\mathbf{r}$ defined on $\Omega_h$ as input and produces a global correction vector $\mathbf{z}$. This is done by combining the results of solving the coarse problem through LU decomposition (bottom) with those of solving all local problems using $\dsstheta$ (top). Note, we use $\Omega_{h,i}$ instead of $R_iAR_i^T$ for reasons detailed in \ref{subsec:archi}.}
  \label{fig:ddml_dss}
\end{figure}

Fig.\ref{fig:ddml_dss} zooms in on the \name{} preconditioner from Fig.\ref{fig:hybrid_solver}, illustrating the application of \name{} to a global residual vector $\mathbf{r}$ within the PCG algorithm.\\

\subsection{Architecture of the $\dsstheta$ model}
\label{subsec:archi}
In addition to our primary contribution, we present a modified version of the original DSS architecture that enables independent inference of solutions regardless of the discretization scheme. In the original DSS approach \cite{deep_statistical_solver}, edge attributes in the MPNNs are derived from the coefficients of $A$. As discussed in Section \ref{subsec:dss}, since $A$ serves as the adjacency matrix for $\Omega_h$, it naturally encodes mesh information. To adapt this architecture, we only need to ensure that the graph is undirected everywhere, except for boundary nodes whose edges point toward the interior of the graph. Thus, we propose utilizing the distance between nodes as edge attributes.
Although computing the linear system remains necessary for training the model, it is no longer required for inferring solutions, thus saving the computational time of generating the linear system. As a consequence, the graph used to infer solutions, as defined in equation \eqref{graph_solutions}, can be equivalently formulated as:

\begin{equation}
G_i = \left(\Omega_{h,i},~\frac{R_i\mathbf{r}}{||R_i\mathbf{r}||}\right)
\end{equation}

\begin{figure}[tb]
  \includegraphics[width=\columnwidth]{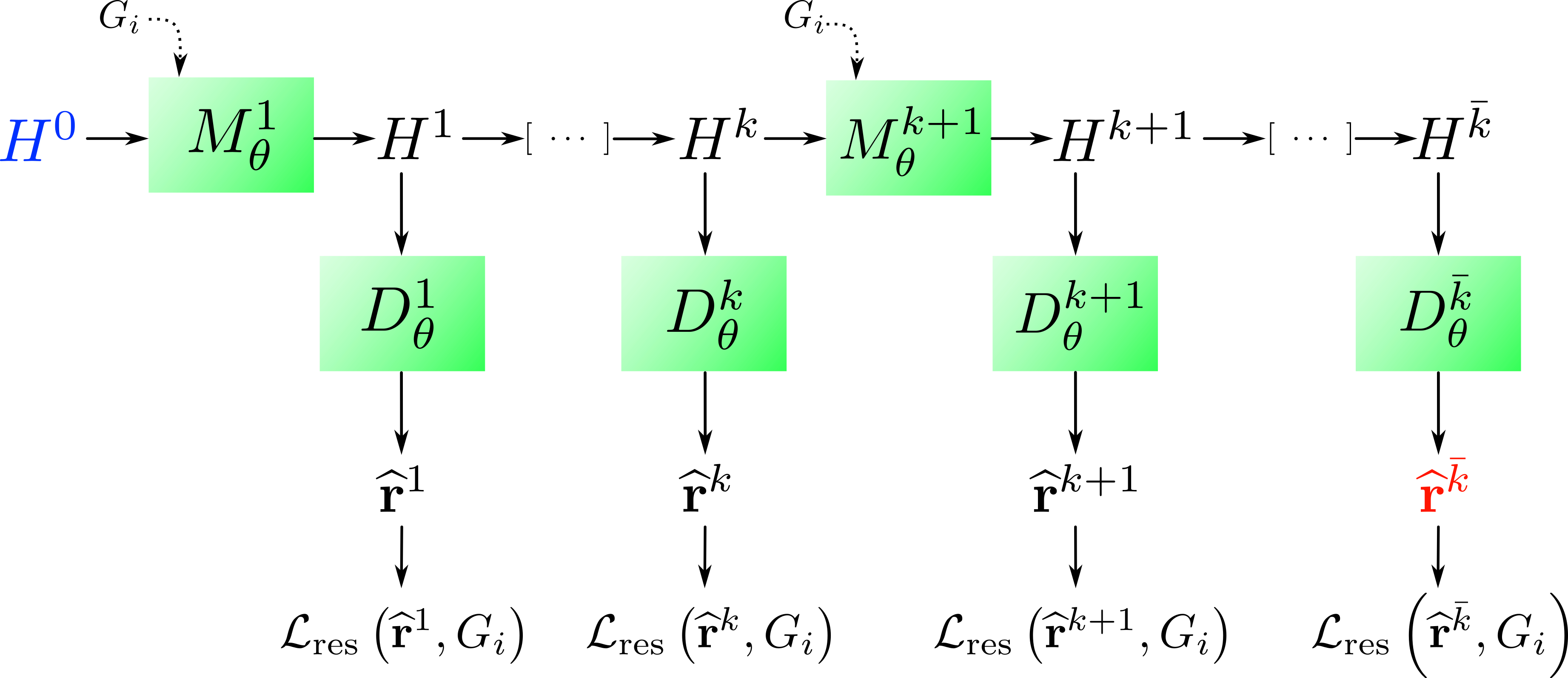}
  \caption{$\dsstheta$ architecture: The model is initialized with a null latent state $H^0$ (in blue). An iterative process then propagates information for $\bar{k}$ iterations using $M^{\bar{k}}_\theta$ distinct blocks of MPNNs. At each iteration $k$, a Decoder $D_\theta^k$ translates the latent state $H^k$ into a physical state $\widehat{\mathbf{r}}^k$, with an intermediate loss computed. The green squares denote trainable functions, and their weights are learned during training. The last $\widehat{\mathbf{r}}^{\bar{k}}$ (in red) represents the model output.}
  \label{fig:dss_archi}
\end{figure}

The architecture of $\dsstheta$, displayed in Fig.\ref{fig:dss_archi}, consists of an iterative process that acts on a latent state $H = (\mathbf{h}_j)_{j \in [k_i]} \in \mathcal{H}$ with $\mathbf{h}_j \in \mathbb{R}^d$, $d \geq 1$ for $\bar{k}$ iterations. The entire architecture can be divided into three steps: an Initialization step, a Message Passing step, and a Decoding step, described as follows. \\

\textbf{1. Initialization} \quad All latent states in $H^0$ are initialized as null vectors. The latent state $H$ represents an embedding of the physical state $\mathbf{r}$ into a higher-dimensional space. This architectural choice is driven by the need to provide enough space for information propagation throughout the network. \\

\textbf{2. Message Passing} \quad The Message Passing step, responsible for the flow of information within the graph, performs $\bar{k}$ updates on the latent state variable $H$ using $\bar{k}$ updating blocks of neural networks. To achieve this, at each iteration $k$, two different messages $\bphi_{\rightarrow}^k$ and $\bphi_{\leftarrow}^k$ are first computed using multi-layer perceptrons (MLPs) $\Phi_{\rightarrow,\theta}^{k+1}$ and $\Phi_{\leftarrow,\theta}^{k+1}$. These messages correspond to outgoing and ingoing links and are defined, node-wise, as follows: 

\begin{align}
\displaystyle \bphi_{\rightarrow,j}^{k+1} &= \sum_{l\in\mathcal{N}(j)} \Phi_{\rightarrow,\theta}^{k+1}(\mathbf{h}_j^k, \mathbf{h}_l^k, \mathbf{d}_{jl}, \norm{\mathbf{d}_{jl}})
\label{dss_mp_out} \\
\displaystyle \bphi_{\leftarrow,j}^{k+1} &= \sum_{l\in\mathcal{N}(j)} \Phi_{\leftarrow,\theta}^{k+1}(\mathbf{h}_j^k, \mathbf{h}_l^k, \mathbf{d}_{lj}, \norm{\mathbf{d}_{lj}})
\label{dss_mp_in}
\end{align}

where $\mathbf{d}_{jl}$ represents the relative position vector and $\norm{\mathbf{d}_{jl}}$ its Euclidean distance. The updated latent state $H^{k+1}$ is then computed using an MLP $\Psi_\theta^{k+1}$ in a ResNet-like fashion such that, node-wise:

\begin{equation}
    \displaystyle \mathbf{h}_j^{k+1} = \mathbf{h}_j^k + \alpha ~ \Psi_\theta^{k+1}(\mathbf{h}_j^k, c_j, \bphi_{\rightarrow,j}^{k+1}, \bphi_{\leftarrow,j}^{k+1})
    \label{dss_resnet}
\end{equation}

with $c = \frac{R_i\mathbf{r}}{||R_i\mathbf{r}||}$. For purpose of clarity, operations \eqref{dss_mp_out}, \eqref{dss_mp_in} and \eqref{dss_resnet} can be grouped into a single updating block $M^{k+1}_\theta$ such that the next latent state $H^{k+1}$ is computed as follows:

\begin{equation}
    H^{k+1} = M^{k+1}_\theta(H^{k},~G_i)
    \label{eq:dss_block}
\end{equation}

\textbf{3. Decoder} \quad A Decoding step is applied after each iteration to convert the latent state $H^{k+1}$ into a meaningful actual state $\widehat{\mathbf{r}}^{k+1}$ by using an MLP $D^{k+1}_\theta$ such that : 

\begin{equation}
    \displaystyle \widehat{\mathbf{r}}^{k+1} = D_\theta^{k+1}(H^{k+1})
    \label{eq:dss_decoder}
\end{equation}

The final state $\widehat{\mathbf{r}}^{\bar{k}}$ represents the actual output of the algorithm, which corresponds to the sought approximate solution $\mathbf{\tilde{r}}_i$ of a local problem $i$ in \eqref{eq:dss_asm_gnn}. The training loss is computed as a sum of all intermediate losses, as follows:

\begin{equation}
    \text{Training Loss } = \displaystyle \sum_{k = 1}^{\bar{k}} \mathcal{L}_\text{res}(\widehat{\mathbf{r}}^k,~G_i)  
    \label{eq:dss_loss}
\end{equation}

\section{Results}
\label{sec:results}

This section analyzes the numerical behavior and performance of the proposed hybrid solver. Section \ref{subsec:dataset} introduces the dataset and Section \ref{subsec:model_config} the training configuration of $\dsstheta$. The numerical behavior of the approach is evaluated in Section \ref{subsec:numerical_behvaior}, and the impact of $\dsstheta$ hyperparameters on performance is assessed in Section \ref{subsec:impact_features_performance}. Finally, the proposed method is benchmarked against an optimized C++ legacy linear solver in Section \ref{subsec:performance_benchmark}.

\subsection{Dataset}
\label{subsec:dataset}

\begin{figure}[tb]
     \centering
     \begin{subfigure}[b]{0.49\columnwidth}
         \centering
         \includegraphics[width=0.85\columnwidth]{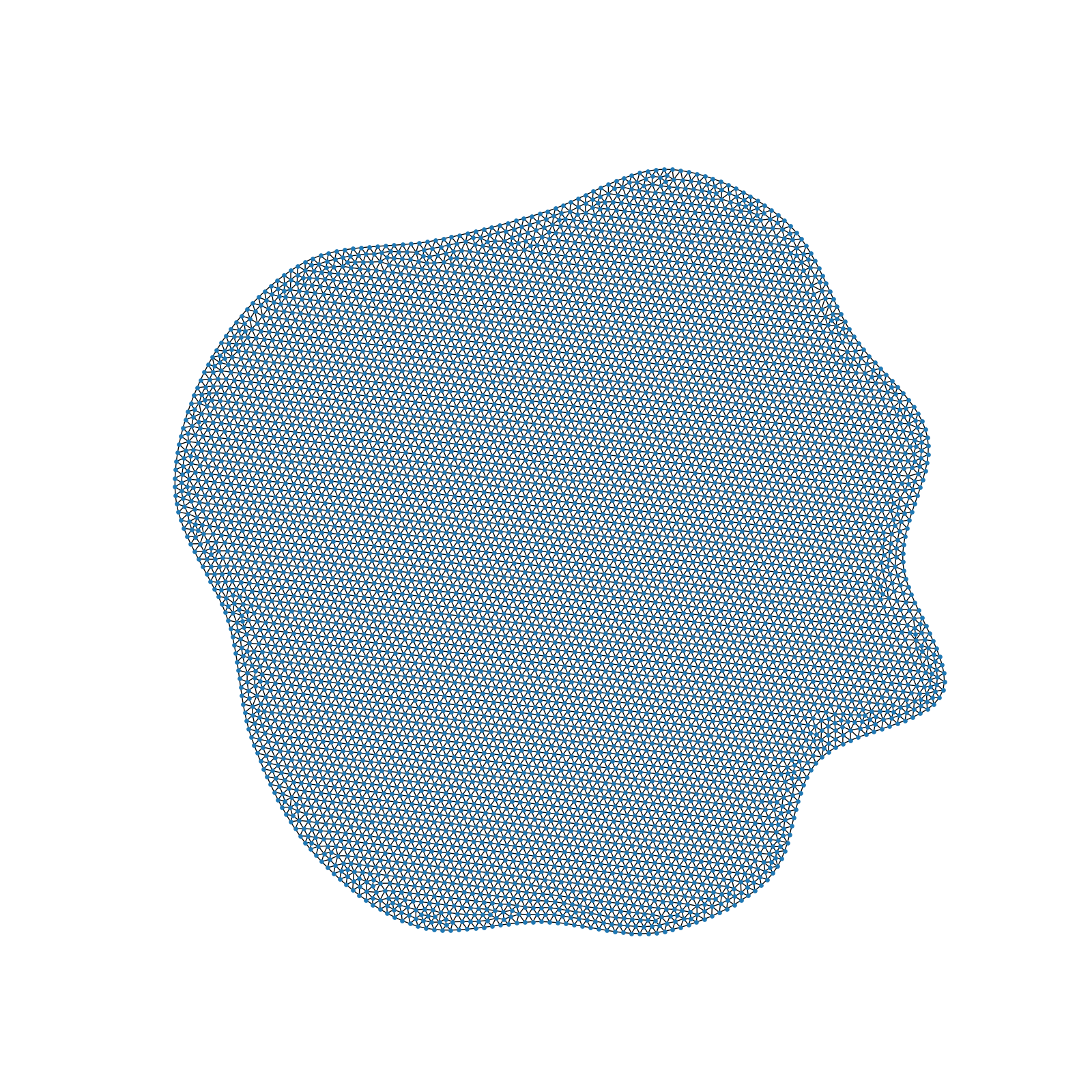}
         \caption{}
         \label{subfig:dataset_graph}
     \end{subfigure}
     \hfill
     \begin{subfigure}[b]{0.49\columnwidth}
         \centering
         \includegraphics[width=0.85\columnwidth]{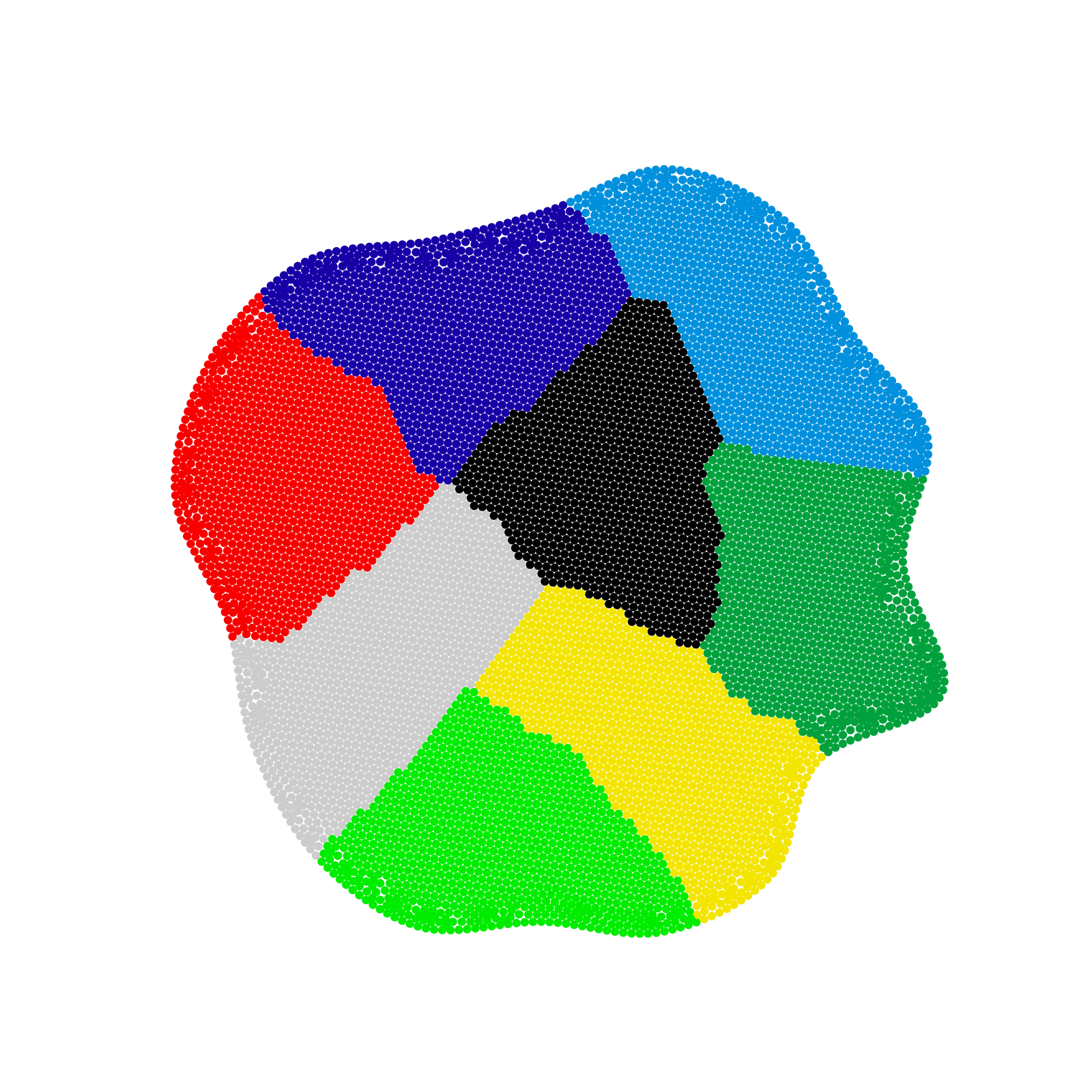}
         \caption{}
         \label{subfig:dataset_partitioning}
     \end{subfigure}
     \caption{Example of (\ref{subfig:dataset_graph}): one generated global domain with $7420$ nodes, and ($\ref{subfig:dataset_partitioning}$): its partitioning into $8$ sub-meshes. Each colored sub-mesh corresponds to a graph sample of the dataset.}
     \label{fig:dataset}
\end{figure}

We generate random $2$D domains $\Omega$ using $20$ points sampled from the unit sphere, connected with Bezier curves to form $\partial \Omega$. Next, GMSH\footnote{\url{https://gmsh.info/}} is used to discretize $\Omega$ into unstructured triangular meshes $\Omega_h$, each containing $6000$ to $8000$ nodes. For each domain, we solve Poisson problems \eqref{poisson_equation} with forcing functions $f$ and boundary functions $g$ defined as random quadratic polynomials with coefficients $r_{i \in \left[1, \cdots, 9\right]}$ uniformly sampled in $[-10,10]$, such that:
\begin{align}
f(x,y) & = r_1(x-1)^2 + r_2 y^2 + r_3 \\
g(x,y) & = r_4x^2 + r_5y^2 + r_6xy + r_7x + r_8y + r_9 
\end{align}
Each mesh $\Omega_h$ is then partitioned into sub-meshes of approximately $1000$ nodes using METIS\footnote{\url{https://github.com/KarypisLab/METIS}} with an overlap of $2$.
We use the PCG solver described in Algorithm \ref{algo:pcg} preconditioned with a classic two-level ASM method (see Section \ref{subsec:asm}) to solve the global Poisson problems with a relative residual norm tolerance of $10^{-6}$. The dataset consists of discretized local Poisson problems from the two-level ASM preconditioner extracted at each iteration of the PCG algorithm. In that setting, solving $500$ global Poisson problems results in $117,138$ samples, further split into $70,282$ / $23,428$ / $23,428$ training/validation/test datasets.
Figure \ref{fig:dataset} displays a global problem and its partitioning into several sub-meshes, each of which belongs to the dataset. In subsequent analyses, when considering meshes with varying node counts, this is achieved by increasing the radius of the mesh while maintaining the element size fixed from the original distribution. The force and boundary functions are rescaled accordingly.

\subsection{Model Configuration \& Training}
\label{subsec:model_config}

$\dsstheta$ is implemented in Pytorch, using the library Pytorch Geometric\footnote{\url{https://github.com/pyg-team/pytorch_geometric}}. The model is trained using the original hyperparameters from \cite{deep_statistical_solver}; the number of iterations $\bar{k}$ is set to $30$, and the latent space dimension $d$ to $10$. Each neural network in Eqs. \eqref{dss_mp_in}, \eqref{dss_mp_out}, \eqref{dss_resnet}, and \eqref{eq:dss_decoder} has one hidden layer of dimension $10$ with a ReLU activation function. All model parameters are initialized using Xavier initialization. In Eq. \eqref{dss_resnet}, coefficient $\alpha$ is set to $10^{-3}$. Training is performed for $400$ epochs on 2 P$100$ GPU using the Adam optimizer with a learning rate of $10^{-2}$ and a batch size of $100$. Gradient clipping is used to prevent exploding gradient issues and is set to $10^{-2}$. The \textit{ReduceLrOnPlateau} scheduler from PyTorch is applied, reducing the learning rate by a factor of $0.1$ during training. After training, the evaluation of $\dsstheta$ on the test set results in a residual norm of $0.0058 \pm 0.002$ and a Relative Error with an ``exact'' solution computed through LU decomposition of $0.13 \pm 0.2$. 

\subsection{Numerical behavior of the hybrid solver}
\label{subsec:numerical_behvaior}

\begin{table}[b]
\renewcommand{\arraystretch}{1.5}
\caption{Numerical Behaviour}
\begin{center}
\resizebox{\columnwidth}{!}{
\begin{tabular}{|cccc|ccc|}
\hline
\multicolumn{4}{|c|}{Configuration} & \multicolumn{3}{c|}{Iterations} \\ \hline
\multicolumn{1}{|c|}{$N$}       & \multicolumn{1}{c|}{$N_s$}   & \multicolumn{1}{c|}{K}      & Overlap   & \multicolumn{1}{c|}{\name{}}     & \multicolumn{1}{c|}{\ddmlu{}}       & CG 
\\ \hline
\multicolumn{1}{|c|}{$2632 \pm 90$}     & \multicolumn{1}{c|}{$1000$}       & \multicolumn{1}{c|}{$3 \pm 0$}    & $2$       & \multicolumn{1}{c|}{$22 \pm 1$}   & \multicolumn{1}{c|}{$19 \pm 1$}   & $113 \pm 3$ 
\\ \hline
\multicolumn{1}{|c|}{--}                & \multicolumn{1}{c|}{--}           & \multicolumn{1}{c|}{--}           & $4$       & \multicolumn{1}{c|}{$17 \pm 1$}   & \multicolumn{1}{c|}{$14 \pm 1$}   & -- 
\\ \hline
\multicolumn{1}{|c|}{--}                & \multicolumn{1}{c|}{$500$}        & \multicolumn{1}{c|}{$6 \pm 0$}    & $2$       & \multicolumn{1}{c|}{$24 \pm 1$}   & \multicolumn{1}{c|}{$22 \pm 1$}   & -- 
\\ \hline
\multicolumn{1}{|c|}{--}                & \multicolumn{1}{c|}{$2000$}       & \multicolumn{1}{c|}{$2 \pm 0$}    & $2$       & \multicolumn{1}{c|}{$18 \pm 1$}   & \multicolumn{1}{c|}{$16 \pm 1$}   & -- 
\\ \hline
\multicolumn{1}{|c|}{$7148 \pm 247$}    & \multicolumn{1}{c|}{$1000$}       & \multicolumn{1}{c|}{$7 \pm 1$}    & $2$       & \multicolumn{1}{c|}{$32 \pm 2$}   & \multicolumn{1}{c|}{$28 \pm 2$}   & $183 \pm 6$ 
\\ \hline
\multicolumn{1}{|c|}{--}                & \multicolumn{1}{c|}{--}           & \multicolumn{1}{c|}{-}            & $4$       & \multicolumn{1}{c|}{$27 \pm 2$}   & \multicolumn{1}{c|}{$21 \pm 2$}   & -- 
\\ \hline
\multicolumn{1}{|c|}{--}                & \multicolumn{1}{c|}{$500$}        & \multicolumn{1}{c|}{$14 \pm 1$}   & $2$       & \multicolumn{1}{c|}{$31 \pm 2$}   & \multicolumn{1}{c|}{$29 \pm 2$}   & -- 
\\ \hline
\multicolumn{1}{|c|}{--}                & \multicolumn{1}{c|}{$2000$}       & \multicolumn{1}{c|}{$4 \pm 0$}    & $2$       & \multicolumn{1}{c|}{$29 \pm 2$}   & \multicolumn{1}{c|}{$24 \pm 2$}   & -- 
\\ \hline
\multicolumn{1}{|c|}{$33969 \pm 1181$}  & \multicolumn{1}{c|}{$1000$}       & \multicolumn{1}{c|}{$35 \pm 2$}   & $2$       & \multicolumn{1}{c|}{$49 \pm 3$}   & \multicolumn{1}{c|}{$42 \pm 3$}   & $385 \pm 6$ 
\\ \hline
\multicolumn{1}{|c|}{--}                & \multicolumn{1}{c|}{--}           & \multicolumn{1}{c|}{--}           & $4$       & \multicolumn{1}{c|}{$46 \pm 3$}   & \multicolumn{1}{c|}{$36 \pm 3$}   & -- 
\\ \hline
\multicolumn{1}{|c|}{--}                & \multicolumn{1}{c|}{$500$}        & \multicolumn{1}{c|}{$68 \pm 2$}   & $2$       & \multicolumn{1}{c|}{$45 \pm 3$}   & \multicolumn{1}{c|}{$42 \pm 3$}   & -- 
\\ \hline
\multicolumn{1}{|c|}{--}                & \multicolumn{1}{c|}{$2000$}       & \multicolumn{1}{c|}{$17 \pm 1$}   & $2$       & \multicolumn{1}{c|}{$51 \pm 3$}   & \multicolumn{1}{c|}{$41 \pm 2$}   & -- 
\\ \hline
\end{tabular}
}
\label{tab:numerical_behavior}
\end{center}
\end{table}
This section examines the numerical behavior of the hybrid solver across various configurations, including out-of-distribution problems. For each setup, we solve $100$ global Poisson problems, sampling parameters as described in Section \ref{subsec:dataset}. The iteration count required to achieve a relative residual norm of $10^{-6}$ is evaluated for different problem sizes $N \simeq 2000$, $7000$ (training configuration), and $30,000$. For each problem size $N$, we test sub-mesh sizes $N_s = 500$, $1000$ (training configuration), and $2000$, with an overlap of $2$. We also investigate local problems of size $1000$ with an overlap of $4$. Table \ref{tab:numerical_behavior} reports mean ($\pm$ standard deviation) values over the $100$ Poisson problems. The proposed approach, PCG preconditioned with \name{}, is compared with PCG preconditioned with \ddmlu{} (i.e. two-level ASM from Section \ref{subsec:asm} with local problems solved through LU decomposition), and a Conjugate Gradient (CG) without preconditioner. Results show that PCG-\ddmlu{} always converges in fewer iterations than PCG-\name{}. This entirely expected behavior is due to the fact that \ddmlu{} uses a direct solver to solve local sub-problems while \name{} uses $\dsstheta$, which cannot achieve higher precision than a direct LU solver. What is interesting is analyzing the convergence of PCG-\name{} in comparison to PCG-\ddmlu{}. Firstly, regardless of the problem or solver setup, PCG-\name{} always converges to the desired precision. Secondly, the difference in iteration counts between PCG-\name{} and PCG-\ddmlu{} is minimal (always fewer than $10$ iterations) across all configurations. It's worth noting that $\dsstheta$ is trained on datasets with $1000$ nodes, yet the entire process consistently converges, even when the sub-mesh size varies (either smaller at $500$ or larger at $2000$), indicating robustness to out-of-distribution samples. Moreover, the proposed approach maintains the advantageous property of faster convergence with larger overlaps. Overall, the proposed hybrid solver demonstrates consistency; leveraging a GNN-based solver does not significantly degrade results compared to using an ``exact solver''. Additionally, the method proves to be scalable in terms of the number of sub-domains and converges much more effectively than CG. \\
\begin{figure}[tb]
     \centering
     \begin{subfigure}[b]{\columnwidth}
         \hspace{1.1cm}
         \includegraphics[width=0.85\columnwidth]{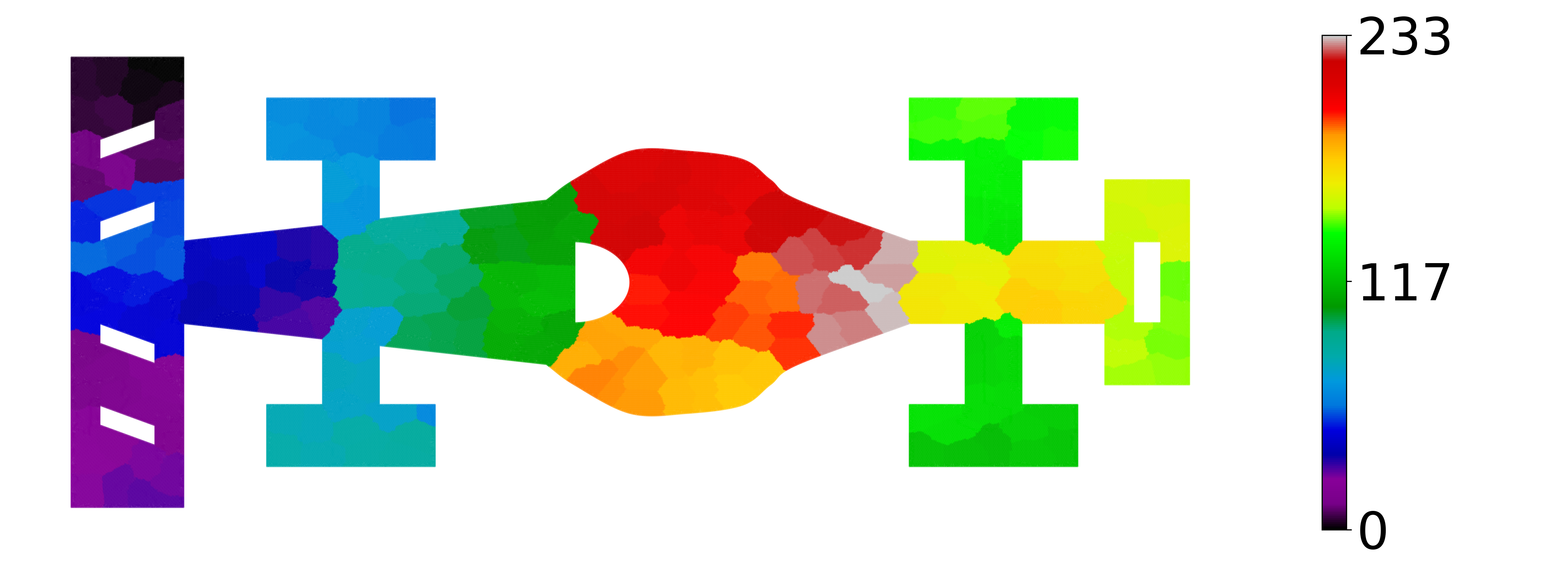}
         \caption{}
         \label{subfig:partitioning}
     \end{subfigure}
     \hfill
     \begin{subfigure}[b]{\columnwidth}
         \centering
         \includegraphics[width=0.9\columnwidth]{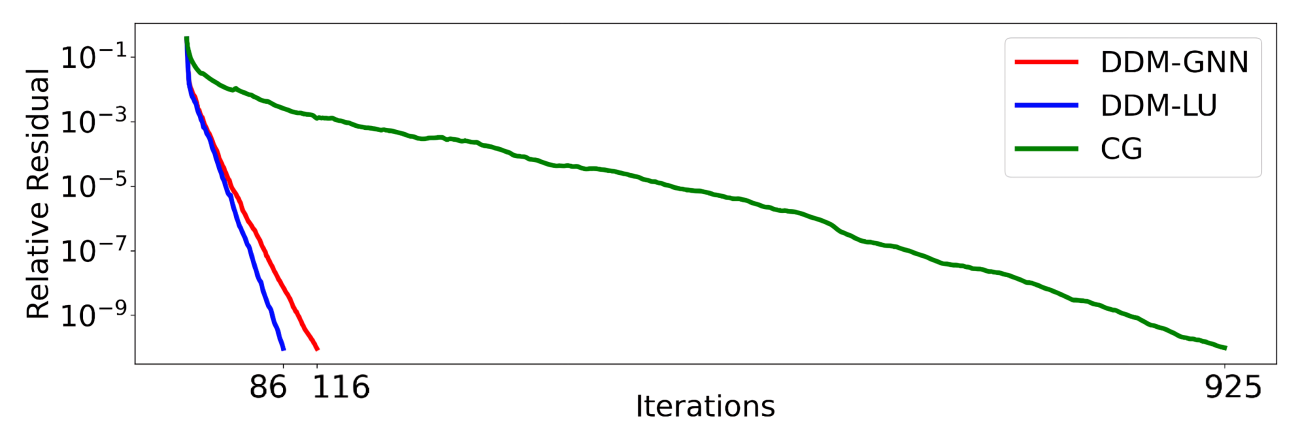}
         \caption{}
         \label{subfig:residual}
     \end{subfigure}
     \caption{(\ref{subfig:partitioning}): partitioning of the mesh into $234$ sub-meshes and ($\ref{subfig:residual}$): the evolution of the relative residual error for the different methods until it reaches an order of $ 10^{-9}$.}
     \label{fig:f1sample}
\end{figure}

We further conduct a large-scale experiment on a mesh representing a caricatural Formula $1$ with $233,246$ nodes. This mesh includes ``holes'' (such as a cockpit and front and rear wing stripes) and is larger than those seen in the training dataset, providing a challenging test of the ability of the hybrid solver to generalize to out-of-distribution (with respect to the geometry of the domain, as well as the mesh size) examples. To solve it, we sample force and boundary functions as described in Section \ref{subsec:dataset}. The mesh is divided into sub-meshes of about $1000$ nodes, resulting in $234$ sub-meshes, as shown in Fig.\ref{subfig:partitioning}. We solve the Poisson problem using PCG preconditioned with \name{} and \ddmlu{}, as well as CG until the relative residual error reaches $10^{-9}$. Figure \ref{subfig:residual} illustrates the evolution of the relative residual error during the iterations of the various methods. Results show that the proposed hybrid solver still converges when applied to an out-of-distribution and large-scale sample, achieving precision levels significantly lower than those used for training the dataset. Besides, PCG-\name{} is still competitive with respect to PCG-\ddmlu{} regarding the iteration count and converges much more efficiently than a traditional Conjugate Gradient. 

\subsection{Impact of $\dsstheta$ hyperparameters on the performance}
\label{subsec:impact_features_performance}

The size of the $\dsstheta$ network is directly influenced by the latent space dimension $d$ and the number of layers in the network $\bar{k}$. From a performance perspective, it is essential to consider that a larger network may result in a more precise trained model, but it also incurs increased inference costs. This exploration analyses the impact of the two parameters, $d$ and $\bar{k}$, on both the numerical behavior (iteration counts at convergence) and the time performance of the hybrid solver. 

Several $\dsstheta$ models are trained corresponding to parameter sets $\bar{k} \in \{5, 10, 20, 30\}$ and $d \in \{5, 10, 20\}$. Table \ref{tab:metrics_dss} summarizes the metrics averaged over the whole test set for these models, including the residual and relative errors, as well as the number of weights. These results show that as $\bar{k}$ and $d$ increase, the metrics improve, but the number of weights also increases.\\
\begin{table}[tb]
\renewcommand{\arraystretch}{1.5}
\caption{Metrics for $\dsstheta$ trained with varying $\bar{k}$ and $d$.}
\begin{center}
\begin{tabular}{|cc|ccc|}
\hline
\multicolumn{2}{|c|}{$\dsstheta$} & \multicolumn{3}{c|}{Metrics}                                                                 \\ \hline
\multicolumn{1}{|c|}{$\bar{k}$}      & $d$      & \multicolumn{1}{c|}{Residual ($10^{-2}$)} & \multicolumn{1}{c|}{Relative Error} & Nb Weights \\ \hline
\multicolumn{1}{|c|}{5}      & 5      & \multicolumn{1}{c|}{$4.11 \pm 1.2$}           & \multicolumn{1}{c|}{$0.67 \pm 0.1$}            & $1755$               \\ \hline
\multicolumn{1}{|c|}{5}      & 10     & \multicolumn{1}{c|}{$3.84 \pm 1.1$}           & \multicolumn{1}{c|}{$0.66 \pm 0.1$}            & $6255$               \\ \hline
\multicolumn{1}{|c|}{5}      & 20     & \multicolumn{1}{c|}{$3.74 \pm 1.1$}           & \multicolumn{1}{c|}{$0.65 \pm 0.1$}            & $23505$               \\ \hline
\multicolumn{1}{|c|}{10}     & 5      & \multicolumn{1}{c|}{$2.02 \pm 0.7$}           & \multicolumn{1}{c|}{$0.48 \pm 0.4$}            & $3510$               \\ \hline
\multicolumn{1}{|c|}{10}     & 10     & \multicolumn{1}{c|}{$1.58 \pm 0.49$}          & \multicolumn{1}{c|}{$0.44 \pm 0.2$}            & $12510$               \\ \hline
\multicolumn{1}{|c|}{10}     & 20     & \multicolumn{1}{c|}{$1.55 \pm 0.49$}          & \multicolumn{1}{c|}{$0.43 \pm 0.1$}            & $47010$               \\ \hline
\multicolumn{1}{|c|}{20}     & 5      & \multicolumn{1}{c|}{$0.88 \pm 0.34$}          & \multicolumn{1}{c|}{$0.20 \pm 0.4$}            & $7020$               \\ \hline
\multicolumn{1}{|c|}{20}     & 10     & \multicolumn{1}{c|}{$0.68 \pm 0.25$}          & \multicolumn{1}{c|}{$0.20 \pm 0.3$}            & $25020$               \\ \hline
\multicolumn{1}{|c|}{20}     & 20     & \multicolumn{1}{c|}{$0.57 \pm 0.21$}          & \multicolumn{1}{c|}{$0.18 \pm 0.1$}            & $94020$               \\ \hline
\multicolumn{1}{|c|}{30}     & 10     & \multicolumn{1}{c|}{$0.58 \pm 0.22$}          & \multicolumn{1}{c|}{$0.13 \pm 0.2$}            & $37530$               \\ \hline
\end{tabular}
\label{tab:metrics_dss}
\end{center}
\end{table}
\begin{figure}[b]
     \centering
     \begin{subfigure}[b]{0.49\columnwidth}
         \centering
         \includegraphics[width=\columnwidth]{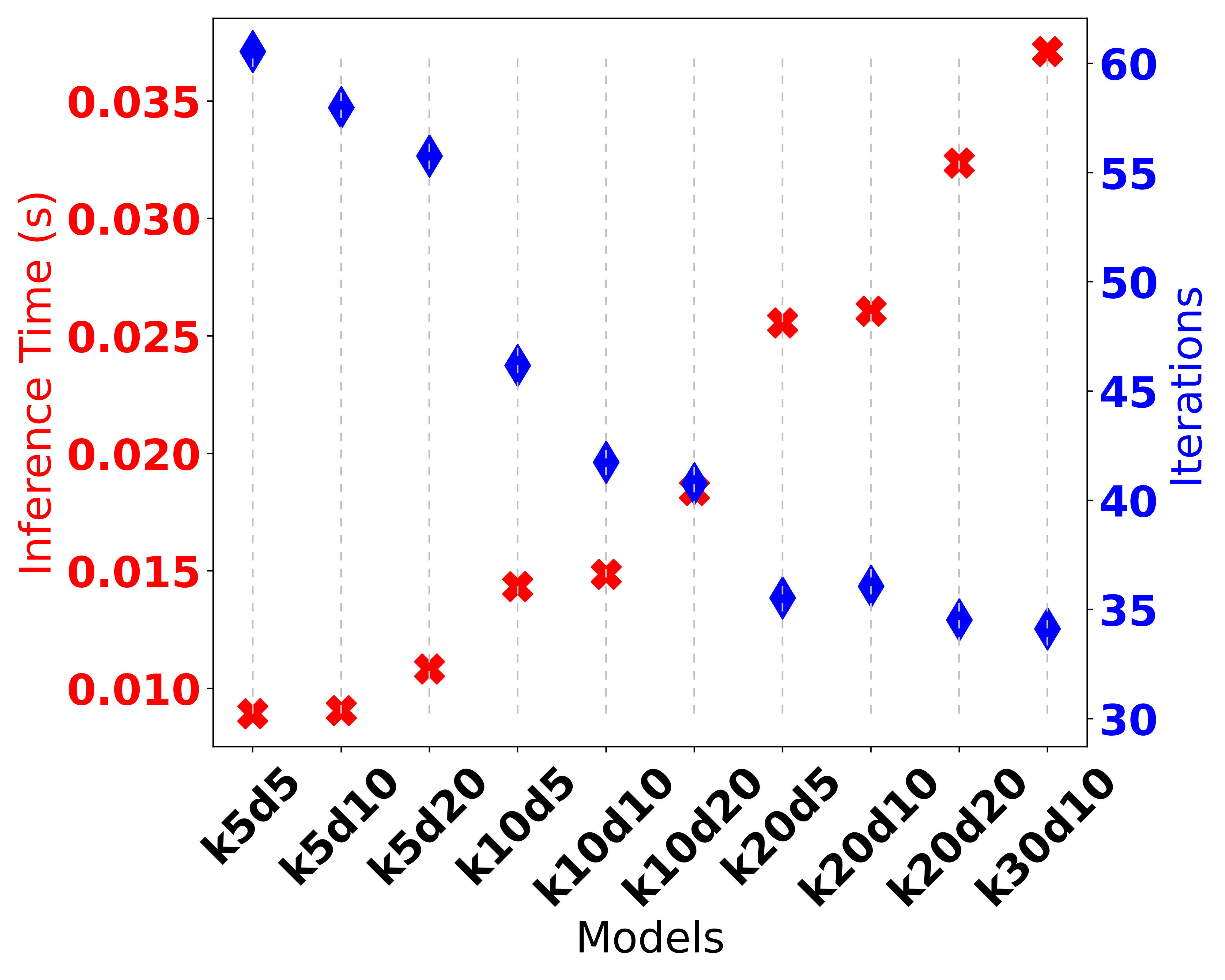}
         \caption{}
         \label{subfig:inference_time}
     \end{subfigure}
     \hfill
     \begin{subfigure}[b]{0.49\columnwidth}
         \centering
         \includegraphics[width=\columnwidth]{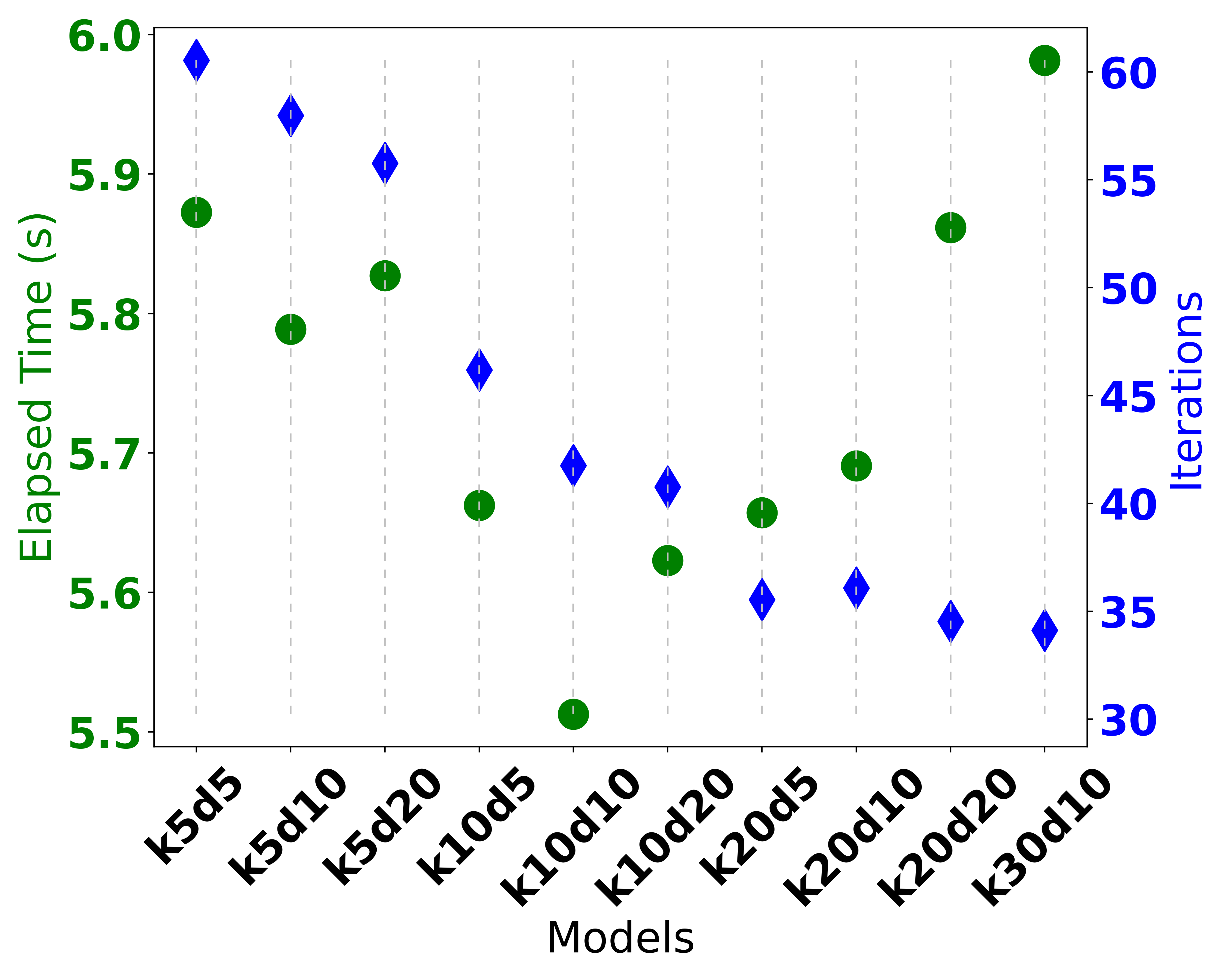}
         \caption{}
         \label{subfig:global_time}
     \end{subfigure}
     \caption{$\dsstheta$ performance exploration regarding $\bar{k}$ and $d$.}
     \label{fig:modelperfexplo}
\end{figure}

Next, we apply the PCG-\name{} solver based on each of these trained models to solve $100$ Poisson problems of size $10,000$. Fig.\ref{subfig:inference_time} illustrates the inference time (in seconds) to solve a batch of local problems on the left axis and the number of iterations at convergence on the right axis for each parameter pair ($\bar{k}$,$d$). Fig.\ref{subfig:global_time} illustrates the elapsed time (in seconds) of the full resolution on the left axis and the number of iterations at convergence on the right axis for each parameter pair ($\bar{k}$,$d$). Upon analysis of Fig.\ref{subfig:inference_time}, as expected, it becomes evident that a larger $\dsstheta$ model (represented by the values of $\bar{k}$ and $d$) results in a greater number of weights, making the preconditioner more accurate at the expense of performance. Fig.\ref{subfig:global_time} highlights that the best performance is achieved with the model corresponding to $\bar{k}=10$ and $d=10$, even though this model may not be the most accurate.

\subsection{Performance results}
\label{subsec:performance_benchmark}

We integrated our \name{} preconditioner into the \soft{HTSSolver} package \cite{ddml, gratien2020introducing}, an in-house C++ linear solver. This required extending its multi-level Domain Decomposition framework, reliant on the direct sparse LU solver from the Eigen\footnote{\url{https://gitlab.com/libeigen/eigen}} library, by incorporating the $\dsstheta$ model. This new solver involves constructing related graph structures for each sub-domain and inferring $\dsstheta$. Regarding $\dsstheta$ inference, we evaluated several C++ inference engines, including ONNX\footnote{\url{https://github.com/onnx/onnx}}, TensorRT\footnote{\url{https://github.com/NVIDIA/TensorRT}}, and the C++ LibTorch\footnote{\url{https://pytorch.org/cppdocs/}} library. A comparative study of these frameworks is presented in \cite{sever2021performance}. Our GNN model, implemented using Pytorch-Geometric, requires a $64$-bit index type for graph structures, making it incompatible with TensorRT. Ultimately, the C++ LibTorch library was chosen for its superior performance on NVidia GPUs.\\ 

The numerical behavior of \name{} implemented in C++ was initially validated by comparing results with the Python framework. We then explore its extensibility concerning global problem sizes and compare its performance with standard optimized preconditioners and the original \ddmlu{} method, which leverages the sparse LU solver from Eigen, one of the most optimized C++ solver packages. To assess performance in terms of elapsed time, experiments were conducted on Jean Zay, an HPE SGI $8600$ multi-node Linux cluster featuring accelerated dual-socket nodes. These nodes are equipped with Intel Cascade Lake $6248$ processors at $2.5$ GHz, each comprising $20$ cores per socket, and either $4$ Nvidia Tesla V$100$ SXM$2$ GPUs with $16$ or $32$ GB of memory. Additionally, the cluster includes accelerated dual-socket nodes with AMD Milan EPYC $7543$ processors at $2.80$ GHz, featuring $32$ cores per socket and $8$ Nvidia A$100$ SXM$4$ GPUs with $80$ GB of memory.\\

To evaluate performance, we analyze the resolution of Poisson problems on increasingly large meshes, resulting in linear systems with sizes approximately $N \simeq 10$k, $40$k, $100$k, $250$k, $400$k, and $600$k. For each system, the graph is partitioned using METIS into $K$ sub-domains of sizes approximately $500$, $1000$, and $2000$ nodes. All experiments solve the linear systems using PCG with a relative residual norm tolerance of $10^{-3}$. We compare the performances of the Incomplete Cholesky preconditioner IC(0), \ddmlu{}, and \name{}. Table \ref{tab:benchmark} summarizes, for each size $N$ and number of sub-domains $K$, the total execution time $T$ to solve the entire problem, the inference times $T_{lu}$ and $T_{gnn}$ for applying \ddmlu{} and \name{}, respectively, and the number of iterations at convergence $N_{iter}$. \\

The analysis of the results shows that we recover the numerical behavior in terms of the number of iterations at convergence $N_{iter}$ described previously. The variation of $N_{iter}$ is less sensitive to the global size $N$ of the linear system for the \ddmlu{} and \name{} preconditioners compared to IC(0). Regarding $K$, the number of sub-domains, the analysis shows that although the GNN model was trained on meshes of size $1000$, the method can generalize even with sub-domains of sizes varying between $500$ and $2000$ nodes. However, the performance analysis in terms of elapsed time reveals a significant challenge in inferring GNN models in C++ code. The ratios $\frac{T_{lu}}{T}$ and $\frac{T_{gnn}}{T}$ indicate that the majority of the overall resolution time is spent in the GNN and LU resolution. Using the LibTorch inference engine, the GNN resolution remains a substantial challenge, preventing our method from being as competitive as expected compared to state-of-the-art optimized preconditioners. In fact, as described in the literature, for instance, in \cite{lazar2023accelerating}, standard inference engines optimized for Image processing or Large Language Models are not yet mature enough to efficiently support GNN algorithms for HPC software. Future work is necessary to enhance the performance of the GNN resolution by developing an efficient inference engine for GNN structures, contributing to either the ONNX or the TensorRT project.
\begin{table}[tb]
\renewcommand{\arraystretch}{1.5}
\caption{Benchmark to legacy C++ solver}
\begin{center}
\resizebox{\columnwidth}{!}{
\begin{tabular}{|c|c|cc|ccc|ccc|}
\hline
\multicolumn{2}{|c|}{Configuration}&\multicolumn{2}{|c|}{IC(0)}
                                   &\multicolumn{3}{|c|}{\ddmlu{}}
                                   &\multicolumn{3}{|c|}{\name{}}\\
\hline
$N$ & $K$ &$N_{iter}$&$T$ 
                   &$N_{iter}$&$T$&$T_{lu}$ 
                   &$N_{iter}$&$T$&$T_{gnn}$\\
\hline         
10571 & 5 & 35 &0.0127&17&0.049&0.018&29&0.147&0.116 \\ 
      &11 &    &      &14&0.041&0.011&25&0.138&0.099 \\    
      &22 &    &      &12&0.030&0.008&25&0.142&0.098 \\ 
\hline         
41871 &21 &66  &0.0850&18&0.16&0.087 &56&0.55&0.44 \\  
      &42 &    &      &16&0.18&0.064 &33&0.63&0.53 \\  
      &84 &    &      &15&0.15&0.062 &42&0.80&0.68 \\    
\hline         
100307&50 &97  &0.298&66&1.97&1.61 &35&1.25&1.03 \\ 
      &100&    &     &39&1.04&0.76 &35&1.30&1.25 \\
      &200&    &     &54&0.76&0.45 &39&1.54&1.43 \\
\hline  
259604&125&136 &1.21 &33&1.43&1.03 &38&2.77&2.38 \\
      &250&    &     &25&1.07&0.66 &38&2.90&2.39 \\
      &500&    &     &22&1.08&0.51 &36&2.88&2.26 \\
\hline
405344&200&168 &2.39 &35&2.38&1.71 &38&3.60&2.83 \\
      &400&    &     &29&1.92&1.20 &38&3.74&2.84 \\
      &800&    &     &25&1.99&0.90&38&5.64&4.45  \\
\hline
609740&300&221 &4.9  &40&3.13&2.92 &38&5.26&4.13 \\
      &600&    &     &37&3.86&2.33 &39&5.67&5.24 \\
     &1200&    &     &32&2.72&1.73 &39&8.88&6.92  \\
\hline
\end{tabular}
}
\label{tab:benchmark}
 \end{center}
\end{table}

\section{Conclusion and Perspectives}
\label{conclusion}


This paper introduces \name{}, a GNN-based preconditioner that integrates a Graph Neural Network (GNN) model with a Domain Decomposition strategy to enhance the convergence of Krylov methods. This innovative approach allows the GNN model to efficiently handle large-scale meshes in parallel on GPUs. \name{} is also equipped with a two-level method, i.e. a coarse space correction, ensuring the weak scalability of the entire process regarding the number of sub-domains. We validated the numerical behavior of the method on various test cases, demonstrating its applicability and flexibility to a broad range of meshes with varying sizes and shapes. The accuracy of the proposed method ensures the linear solver converges efficiently to any desired level of precision. Implemented within a C++ solver package, the performance of \name{} was evaluated against state-of-the-art optimized preconditioners. However, inferring GNN models in such contexts remains challenging. Existing inference engines like TensorRT or ONNX are not mature enough to handle GNN algorithms efficiently in C++ High-Performance Computing (HPC) codes. Addressing this challenge requires future efforts, including direct contributions to projects like TensorRT or ONNX. To conclude, it is noteworthy that integrating GPU computations into existing legacy codes can be a daunting task. The proposed method holds promise by offering a straightforward methodology that can be easily incorporated into HPC frameworks, harnessing GPU computations. \\

The long-term objective of this research is to accelerate industrial Computation Fluid Dynamics (CFD) codes on software platforms such as OpenFOAM\footnote{\url{https://www.openfoam.com/}}
In upcoming work, we plan to evaluate the efficiency of the proposed approach by implementing it in an industrial CFD code and assessing its impact on performance acceleration.

\section{Acknowledgements}

This research was supported by DATAIA Convergence Institute as part of the “Programmed’Investissement d’Avenir”, (ANR- 17-CONV-0003) operated by INRIA and IFPEN. This project was provided with computer and storage resources by GENCI at [IDRIS] thanks to the grant 2023-[AD011014247] on the supercomputer [Jean Zay]'s the [V100/A100] partition.

\printbibliography

\end{document}